\documentclass[acmtog,authorversion]{acmart}
\acmSubmissionID{313}
\pdfoutput=1

\usepackage{booktabs} 

\setcitestyle{square}

\usepackage[ruled]{algorithm2e} 

\SetAlFnt{\small}
\SetAlCapFnt{\small}
\SetAlCapNameFnt{\small}
\SetAlCapHSkip{0pt}
\IncMargin{-\parindent}

\DeclareMathOperator*{\argmax}{arg\,max}
\newcommand{\nn}{\textrm{NN}}
\newcommand{\etal}{et al.}
\usepackage{color}
\usepackage{soul}
\usepackage[ruled]{algorithm2e} 
\usepackage{rotating}

\definecolor{purple}{rgb}{0.65,0,0.65}
\definecolor{blue}{rgb}{0, 0.2, 0.8}
\definecolor{orange}{rgb}{0.6, 0.6, 0}
\definecolor{red}{rgb}{0.8, 0.2, 0.2}
\definecolor{magenta}{rgb}{0.5, 0.0, 1.0}
\definecolor{black}{rgb}{0.0, 0.0, 0.0}
\definecolor{cyan}{rgb}{0, 0.65, 0.65}
\usepackage{xcolor}

\newif\ifdraft
\draftfalse

\ifdraft
\newcommand{\dlc}[1]{{\color{blue}\textbf{DL:} #1}}
\newcommand{\dcc}[1]{{\color{red}\textbf{DC:} #1}}
\newcommand{\kac}[1]{{\color{orange}\textbf{KA:} #1}}
\newcommand{\nfc}[1]{{\color{green}\textbf{NF:} #1}}

\newcommand{\ka}[1]{{\color{orange}#1}}

\else
\newcommand{\dlc}[1]{}
\newcommand{\dcc}[1]{}
\newcommand{\kac}[1]{}
\newcommand{\nfc}[1]{}

\newcommand{\ka}[1]{{\color{black}#1}}

\fi

\begin{document}
\acmJournal{TOG}
\acmYear{2018}
\acmVolume{37}
\acmNumber{4}
\acmArticle{69}
\acmMonth{8}

\setcopyright{acmcopyright}

\acmDOI{10.1145/3197517.3201332}
\title{Neural Best-Buddies: Sparse Cross-Domain Correspondence}

\author{Kfir Aberman}
\affiliation{%
\institution{AICFVE Beijing Film Academy,}
  \institution{Tel-Aviv University}
}

\author{Jing Liao}
\affiliation{%
  \institution{Microsoft Research Asia (MSRA)}
  }
\author{Mingyi Shi}
\affiliation{%
  \institution{Shandong University}
}  
  \author{Dani Lischinski}
\affiliation{%
  \institution{Hebrew University of Jerusalem}
  }
  \author{Baoquan Chen}
\affiliation{%
  \institution{Shandong University,}
  \institution{AICFVE Beijing Film Academy,}
  \institution{Peking University}
  }
  \author{Daniel Cohen-Or}
\affiliation{%
  \institution{Tel-Aviv University}
}

\renewcommand\shortauthors{Aberman, K. et al}

\begin{abstract}
Correspondence between images is a fundamental problem in computer vision, with a variety of graphics applications. 
This paper presents a novel method for \emph{sparse cross-domain correspondence}.
Our method is designed for pairs of images where the main objects of interest may belong to different semantic categories and differ drastically in shape and appearance, yet still contain semantically related or geometrically similar parts.
Our approach operates on hierarchies of deep features, extracted from the input images by a pre-trained CNN.
Specifically, starting from the coarsest layer in both hierarchies, we search for Neural Best Buddies (NBB): pairs of neurons that are mutual nearest neighbors.
The key idea is then to percolate NBBs through the hierarchy, while narrowing down the search regions at each level and retaining only NBBs with significant activations.
Furthermore, in order to overcome differences in appearance, each pair of search regions is transformed into a common appearance.  

We evaluate our method via a user study, in addition to comparisons with alternative correspondence approaches.
The usefulness of our method is demonstrated using a variety of graphics applications, including cross-domain image alignment, creation of hybrid images, automatic image morphing, and more.
\end{abstract}

%
%

\ccsdesc[300]{Computing methodologies~Interest point and salient region detections}
\ccsdesc[300]{Computing methodologies~Matching}
\ccsdesc[300]{Computing methodologies~Image manipulation}
%
%

\keywords{cross-domain correspondence, image hybrids, image morphing}

\begin{teaserfigure}
\centering
\includegraphics[width=\linewidth]{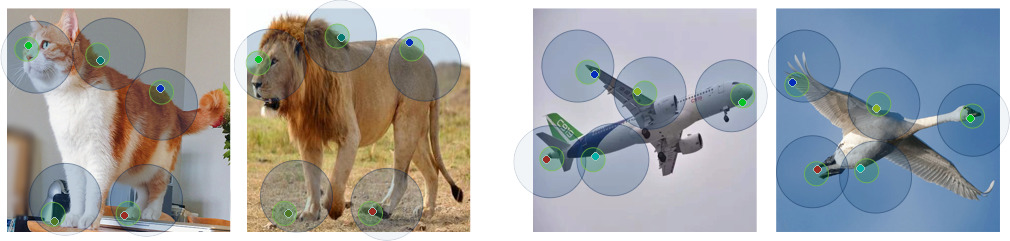} 
\caption{Top $5$ Neural Best-Buddies for two cross-domain image pairs. Using deep features of a pre-trained neural network, our coarse-to-fine sparse correspondence algorithm first finds high-level, low resolution, semantically matching areas (indicated by the large blue circles), then narrows down the search area to intermediate levels (middle green circles), until precise localization on well-defined edges in the pixel space (colored in corresponding unique colors).}
 \label{fig:teaser}
\end{teaserfigure}
 
%




\maketitle

\section{Introduction}
\label{sec:intro}

Finding correspondences between a pair of images has been a long standing problem, with a multitude of applications in computer vision and graphics.
In particular, sparse sets of corresponding point pairs may be used for tasks such as template matching, image alignment, and image morphing, to name a few. 
Over the years, a variety of dense and sparse correspondence methods have been developed, most of which assume that the input images depict the same scene or object (with differences in viewpoint, lighting, object pose, etc.), or a pair of objects from the same class.

In this work, we are concerned with sparse \emph{cross-domain correspondence}: a more general and challenging version of the sparse correspondence problem, where the object of interest in the two input images can differ more drastically in their shape and appearance, such as objects belonging to different semantic categories (domains). It is, however, assumed that the objects contain at least some semantically related parts or geometrically similar regions, otherwise the correspondence task cannot be considered well-defined.
Two examples of cross-domain scenarios and the results of our approach are shown in Figure \ref{fig:teaser}. We focus on sparse correspondence, since in many cross-domain image pairs, dense correspondence is not well-defined; for example, in Figure \ref{fig:teaser}, the lion's mane has no correspondence within the cat image, and no part of the goose corresponds to the plane's engines.


When attempting to find a sparse set of cross-domain correspondences we are faced with two conceptual subproblems:
deciding which points in one image constitute meaningful candidates for matching, and finding their best matching counterparts in the other image. In other words, our goal is to find matching pairs of points, with the requirement that these points are located in important/strategic locations in both images.

Our approach achieves both of these goals in a unified framework \ka{that leverages the power of deep features extractable by a Convolutional Neural Network (CNN), which has been trained for the image classification task.}
Specifically, we adopt the notion of Best Buddies Pairs (BBPs), originally proposed as a similarity measure \cite{dekel2015best}, and extend it to Neural Best Buddies (NBBs), designed to solve the correspondence problem in the challenging cross-domain scenario.

In recent years, CNNs have demonstrated outstanding performance on a variety of vision tasks, including image classification and object detection. It has been shown that the deeper layers of a \ka{trained classification network} extract high-level discriminative features with invariance to position and appearance, while the shallower layers encode low level image features, such as edges and corners, etc. \cite{zeiler2013visualizing,yosinski2015understanding}. Our method leverages on the hierarchical encoding of features by such pre-trained networks. The key idea is to define the correspondence starting from the deeper, semantically meaningful and invariant features.
These correspondences are filtered and their locations are refined, as they are propagated through the layers, until convergence to accurate locations on significant low-level features.

Given two input images, and a trained classification network, two hierarchies of features are built. For each pair of corresponding levels, one from each hierarchy, we extract a sparse set of NBBs. Two neurons are considered best-buddies if they are mutual nearest neighbors, i.e., each neuron is the nearest neighbor of the other in the corresponding set \cite{dekel2015best}. Among the NBBs, we choose to keep only a subset which have high activation values, representing discriminative semantic areas in the deeper layers and key points, such as edges and corners, in the shallower layers.

The spatial positions of the NBBs are refined as they are percolated through the hierarchy, in a coarse-to-fine fashion, by considering only the receptive field of each NBB neuron in the preceding layer of the CNN, until reaching the final positions in the original pair of input images (see the illustration in Figure~\ref{fig:bbps}.)

To enable the computation of the NBBs in a cross-domain setting, the features are first transformed to a common appearance, so that a simple patch correlation could be used effectively to measure point similarity. This is achieved using a simple style transfer technique that modifies the low-order statistics of the features in each pair of regions that we aim to match. 

A variety of graphics applications, such as shape blending and image morphing, require cross-domain correspondences. However, these applications traditionally require manual interaction, leaving the semantic analysis task to the user. We demonstrate a number of such applications, evaluate the performance of our method, and compare it with other state-of-the-art methods. We show a number of fully automated image morphing sequences, created between objects from different semantic categories using our NBBs.
In addition, the uniqueness of our approach is demonstrated via a new image hybridization application, where distinctive parts of two subjects are automatically combined into a hybrid creature, after aligning the two images based on our NBBs.

\section{Related Work}
\label{sec:related}
\subsection{Pairwise Keypoint Matching}
In general, finding a sparse correspondence between two images involves two main steps: extracting individual key points (represented by descriptors), and performing metric-based matching.

There are various techniques to extract key points \cite{harris1988combined}, which are characterized by well-defined positions in the image space, local information content, and stability \cite{lindeberg2015image}. However, in general, keypoint localization for generic object categories remains a challenging task. Most of the existing works on part localization or keypoint prediction focus on either facial landmark localization \cite{belhumeur2013localizing, kowalski2017deep} or human pose estimation \cite{gkioxari2014using}.

In order to identify the extracted points, a local descriptor is generated for each. Local invariant features such as SIFT \cite{lowe2004distinctive}, SURF \cite{bay2006surf} and Daisy \cite{tola2010daisy}, has brought significant progress to a wide range of matching-based applications. These features are robust to typical appearance variations (e.g., illumination, blur) and a wide range of 2D transformations. However, these methods are unable to cope with major dissimilarities between the compared objects, such as strong color and shape differences.

More recently, various CNN-based descriptors were offered to replace traditional gradient based ones \cite{fischer2014descriptor}. Some of the descriptors are used for view-point invariance matching \cite{simonyan2014learning}, others for discriminant patch representations \cite{simo2015discriminative}. Kim \etal~\shortcite{kim2017fcss} offer a descriptor which is based on local self-similarity to robustly match points among different instances within the same object class. 

Ufer \etal~\shortcite{ufer2017deep} suggest a framework for sparse matching by extracting keypoints based on neuron activation, aiming at intra-class cases. However, the fact that the neurons are extracted from one specific layer together with imposed geometric constraints, limits their approach to mainly deal with same class cases, rather than two objects that exhibit a higher level of semantic similarity.

Following the keypoint extraction step, given two sets of descriptors, the matching process is usually performed using nearest neighbor matching, followed by an optional geometric verification. The one-directional matching obtained using the nearest neighbor field can be narrowed down by considering only mutual nearest neighbors. This method was previously leveraged for tasks such as image matching \cite{li2015image}, classification of images, etc., and extended to the Best-Buddies similarity concept, which measures similarity between patches \cite{dekel2015best,talmi2016template} for the purpose of template matching. 

We are not aware of any previous works which aim directly at finding sparse correspondence between two objects belonging to different semantic categories, as we do here.

\subsection{Dense Correspondence}
Normally, sparse matched keypoints constitute a basis for dense correspondence, which is a fundamental tool in applications such as stereo matching, and image registration. For the same scene scenario, a basic densification can be done by assuming a geometric model (affine, homography) that transforms between the two images. However, when the scenes are different, a simple geometric model cannot be assumed.

First steps towards semantic dense correspondence were made by Liu \etal~\shortcite{liu2011sift} with the development of SIFT flow.
In their method, a displacement field is obtained by solving a discrete optimization problem in a hierarchical scheme, based on densely sampled SIFT features.
Following SIFT flow, a number of other flow-based methods were suggested, e.g., Deep flow \cite{weinzaepfel2013deepflow} and Daisy flow \cite{yang2014daisy}, which perform matching of visually different scenes, and \cite{zhou2015flowweb} which proposes a net of correspondences between multiple same-class images, based on cycle-consistent connections.
In parallel to the flow methods that assume smoothness of the flow field, the PatchMatch family \cite{barnes2009patchmatch,barnes2010generalized} relaxes the rigidity assumption, yielding a dense patch-based nearest-neighbor field (NNF) instead.
NRDC \cite{hacohen2011non} extends generalized PatchMatch to cope with significant geometric and photometric variations of the same content.

In addition, CNN-based features (outputs of a certain convolution layer \cite{long2014convnets}, object proposals \cite{ham2016proposal}, etc.) have been employed with flow algorithms, and have shown potential to align intra-class objects better than handcrafted features.
Zhou \etal~\shortcite{zhou2016learning} suggested an end-to-end trained network that requires additional data in the form of synthetic rendered 3D models for formulating a cycle constraint between images.
\ka{Choy \etal~\shortcite{choy2016universal} offered a unified system that learns correspondences based on annotated examples. However, this method is fully supervised,
	and in the case of cross-domain pairs it may be difficult to avoid ambiguities in the annotations.}
Liao \etal~\shortcite{liao2017visual}, combine a coarse-to-fine deep feature pyramid with PatchMatch to compute a semantically-meaningful dense correspondence for transferring
visual attributes between images.
The images may differ in appearance, but must have perceptually similar semantic structure (same type of scene containing objects of similar classes).
The above requirement, along with the goal of computing a dense mapping, makes their approach not well suited for our cross-domain scenario.

\subsection{Image Morphing and Hybridization}
Image morphing is a widely used effect \cite{wolberg1998image}, which typically requires a set of correspondences to define a warp field between the two images.
With images of objects of the same class, such as two human faces, the necessary correspondences can be determined automatically in some cases, e.g., \cite{bichsel1996automatic}.
Morphing between images depicting objects from different classes requires user intervention to provide pairs of corresponding features, or explicit specification of a warp field \cite{liao2014automating}.
Shechtman \etal~\shortcite{shechtman2010regenerative}  
applied patch-based texture synthesis to achieve interesting morph-like
effect in a fully automated way, although this method is automatic it does not involves geometry warps, which make it hard to compare to the classic morph effect.
We are not aware of any general, fully automated method, which is able to provide the correspondences necessary for image morphing between different objects that may even belong to different semantic classes.
Our method enables automated image morphing which is not limited to perform in a specific domain.

Differing from image morphing, 
image hybridization refers to the process of assembling an image from different components from several source images. Hybridization can be performed by seamlessly fusing together parts from different images  \cite{perez2003poisson, huang2013mind}. Some hybridization methods are domain specific, e.g., \cite{bitouk2008face} and \cite{korshunova2016fast}. Hybrid images may also be assembled from different spectral components, such that the image appears to change as the viewing distance changes \cite{oliva2006hybrid}. 
In this work we explore automatic as well as interactive cross-domain image hybridization.

\section{Cross Domain Deep Correspondence}
\label{sec:semanticorrespondence}

\begin{figure}
\centering
\includegraphics[width=0.9\linewidth]{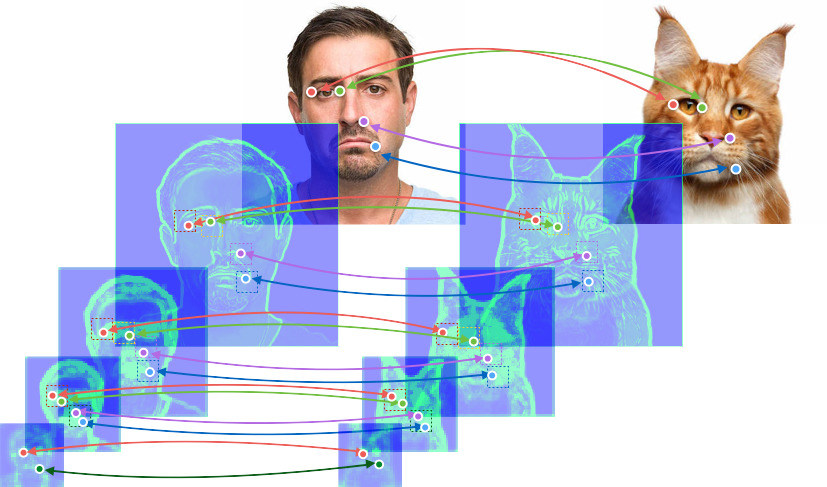}\\
(a)\\
\includegraphics[width=0.9\linewidth]{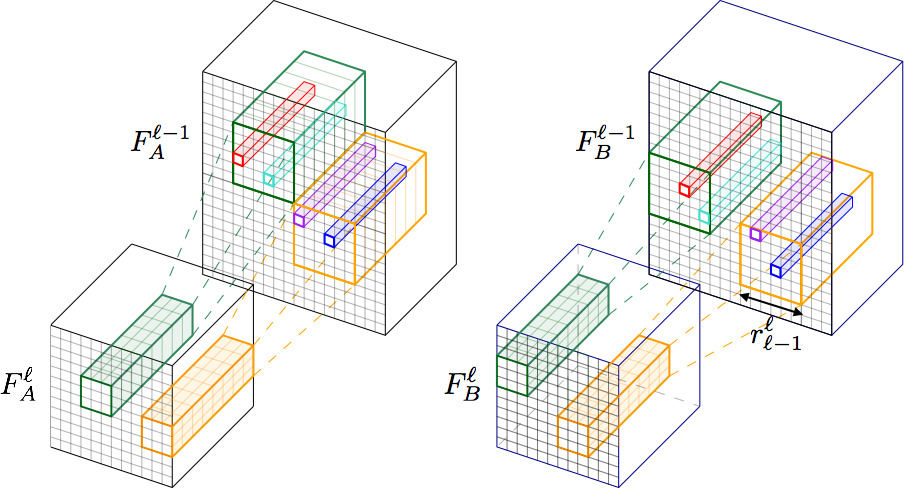}\\
(b)\\
\caption{Sparse semantically meaningful correspondence. (a) At each level, strongly activated NBBs are found in corresponding regions between the two feature maps. (b) The correspondences are propagated to the image pixel level in a coarse-to-fine manner, where at each consecutive finer level, the search area is determined by the receptive fields of the NBBs in the previous layer.}
\label{fig:bbps}
\end{figure}

Given two images whose main regions of interest contain semantically related parts, or geometrically similar patterns, our goal is to find a set of pairwise correspondences.
Furthermore, we strive to find correspondences for points positioned at semantically or geometrically meaningful locations.

Due to the differences in the shape and/or appearance of the objects, our approach is to exploit high-level information.
Such information is encoded by the deep feature maps extracted by CNNs 
\ka{pretrained for the classification task}.
We represent these feature maps as a pyramid, whose top level contains the feature maps extracted by the last convolution layer, and the bottom level consists of the maps from the shallowest layer of the CNN (Section \ref{subsec:deepfeatures}).

Our key idea is to propagate pairs of matching neurons from the top pyramid levels to the bottom ones, while narrowing down the search area at each step (see Figure~\ref{fig:bbps}), and focusing only on meaningful corresponding neurons at each level. 
Furthermore, in order to compensate for differences in appearance, which might be considerable in the cross-domain case, we first transform pairs of corresponding regions to a \emph{common local appearance}, and only then perform a search for matching neurons.
This approach is particularly necessary for finding correct matches within shallow feature maps, which are more strongly correlated with the appearance of the original image than those at deeper layers.

\ka{Note that we do not assume that deep features are invariant across domains. 
Rather, we only assume that corresponding features would be more similar to each other than to others in a small surrounding window.
For example, we expect the features activated by an airplane wing to be more (mutually) similar to those activated by the wing of a goose than to those activated by other nearby parts.}
Thus, we define the corresponding neurons using the notion of meaningful Neural Best Buddies pairs (NBBs). Two neurons are considered best-buddies if they are mutual nearest neighbors, meaning that each neuron is the nearest neighbor of the other, among all the neurons in its set. Rather than keeping all of the NBB pairs, however, we only retain pairs where both neurons are strongly activated. 

Following the extraction of two deep feature pyramids, three main steps are performed at each pyramid level: 
First, we extract NBB candidates from corresponding regions (Section~\ref{subsec:NBBs}).
Second, the NBBs are selected based on the magnitudes of their activations (Section~\ref{subesc:meaningfulBB}).
Third, we propagate the remaining matches into the next hierarchy level, using the receptive fields of the NBB neurons to define refinement search regions around the propagated locations.
Each pair of corresponding regions is transformed to a common appearance, as described in Section~\ref{subesc:commonappear}.
Given the full resulting set of corresponding pairs, we explain in Section~\ref{subsec:kpoints}, how to pick $k$ high quality, spatially scattered correspondences.
The above stages of our algorithm are summarized in Algorithm~\ref{alg:crossdomain}.
 
\subsection{Deep Features Pyramid}
\label{subsec:deepfeatures}

Below, we elaborate on our coarse-to-fine analysis through the feature map hierarchy of a pre-trained network.
Given two input images $I_A$ and $I_B$, they are first fed forward through the VGG-19 network \cite{simonyan2014very}, to yield a five-level feature map pyramid ($\ell = 1,2,3,4,5$), where each level has progressively coarser spatial resolution.
Specifically, the $\ell$-th level is set to the feature tensor produced by the {\tt relu$\ell$\_1} layer of VGG-19. We denote these feature tensors of images $I_A$ and $I_B$ by $F^{\ell}_A$ and $F^{\ell}_B$, respectively.
The feature tensors of the last level capture all the information that enables the subsequent fully connected layers to classify the image into one of the ImageNet categories \cite{russakovsky2015imagenet}.
It has been shown that the deeper layers of a trained network represent larger regions in the original image, encoding higher-level semantic information, which can be used as a descriptor for semantic matching, while the shallower layers encode lower level features over smaller image regions, such as edges, corners, and other simple conjunctions \cite{zeiler2013visualizing}.
We thus exploit the information encoded by the different layers, in order to find and localize semantic correspondences between two objects from different classes, which may differ in overall shape or appearance, but share some semantic similarities.

\subsection{Neural Best Buddies}
\label{subsec:NBBs}

One of the challenges of cross-domain correspondence is that there are parts in one image, or even in the main object of interest, which have no correspondence in the other image/object.
In order to avoid matching these incompatible regions, we utilize the concept of \emph{Neural Best Buddies pairs} (NBBs), which are percolated through the deep features hierarchy, starting from the coarsest level to the finer ones. 
Specifically, at each level, the NBBs are computed only within pairs of corresponding regions defined by the receptive fields of the most meaningful NBBs discovered in the previous (coarser) level in the hierarchy. 
Let $R^{\ell}=\{(P^{\ell}_i, Q^{\ell}_i )\}_{i=1}^{N^{\ell}} $ be a set of $N^\ell$ pairs of corresponding regions in the $\ell$-th layer, where $P^{\ell}_i$ and $Q^{\ell}_i$ are subsets of the spatial domains of $F^{\ell}_A$ and $F^{\ell}_B$, respectively. 
In the top level ($\ell=5$), the set of corresponding regions is initialized to a single pair $R^5 = \{(P^{5}, Q^{5})\}$, representing the entire domain of $F_A^5$ and $F_B^5$.

For each pair of regions, $(P^{\ell}_i, Q^{\ell}_i )\in R^{\ell}$, we extract its NBBs. More formally, a pair of neurons, $(p_j^{\ell}\in P^{\ell}_i ,q_j^{\ell} \in  Q^{\ell}_i)$, is defined as NBB pair if they are mutual nearest neighbors:
\begin{equation}
 \nn^{P^{\ell}_i \rightarrow Q^{\ell}_i} (p_j^{\ell}) =q_j^{\ell} \; \textrm{and} \;\nn^{Q^{\ell}_i \rightarrow P^{\ell}_i} (q_j^{\ell}) =p_j^{\ell},
\end{equation}
where $\nn^{P \rightarrow Q} (p)$ is the Nearest Neighbor of neuron $p\in P$ in the set $Q$ under a similarity metric function $d$:
\begin{equation}
	 \nn^{P \rightarrow Q} (p) = \argmax_{q\in Q}  d(p,q,P,Q).
\end{equation}
Next, we define our deep feature similarity metric, $d(p,q,P,Q)$, to facilitate the matching in a cross-domain setting.

\begin{figure}
\centering
\includegraphics[width=\linewidth]{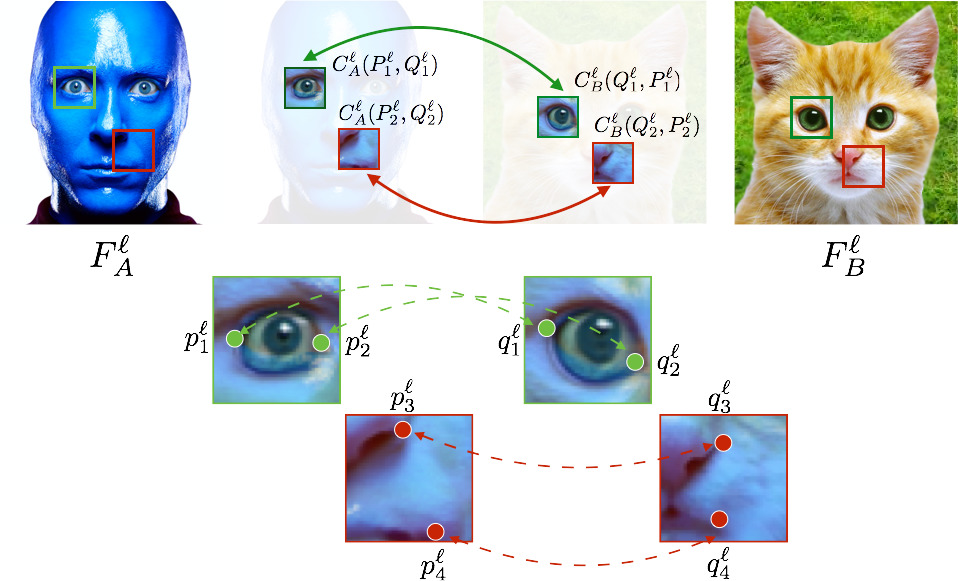}
\caption{Local style transfer to a common appearance. Starting from the original image features, $F_A^{\ell}, F_B^{\ell}$, we transfer the style of each pair of corresponding regions $(P^{\ell}_i,Q^{\ell}_i)$ to their average style, obtaining $C^{\ell}_{A}(P^{\ell}_i, Q^{\ell}_i)$ and $C_{B}^{\ell}(Q^{\ell}_i, P^{\ell}_i)$. A local search is then used to extract NBB pairs $(p^{\ell}_i,q^{\ell}_i)$.
\vspace{1em}
}
\label{fig:midstyle}
\end{figure}

\begin{figure}
\centering
\includegraphics[width=\linewidth]{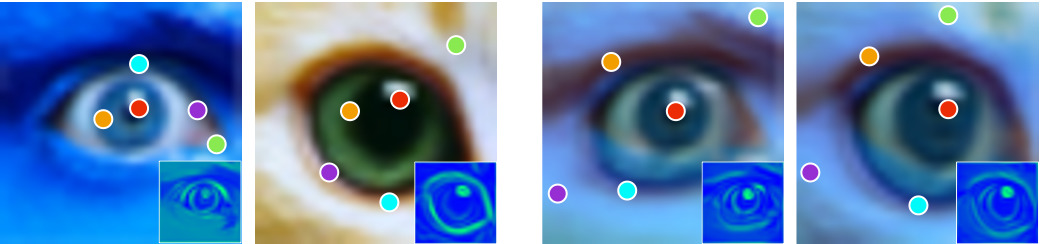}\\
\begin{tabular}{cc}
(a) &  \hspace{3.7cm}(b)
\end{tabular}
\caption{The advantage of common local appearance for low level features ($\ell=1$) patch similarity. (a) NBBs ($k=5$) based on standard patch correlation. (b) NBBs ($k=5$) based on our common appearance metric. Note that the search is performed over the deep features space (bottom right corner), and the points are presented on the reconstructed original image pixels only for demonstration.}
\label{fig:midstyle_heatmap}
\end{figure}

\subsection{Common Local Appearance}
\label{subesc:commonappear}

Naive patch similarity, using a metric such as $L_2$ between deep features, is valid only in the top levels, since lower levels are more strongly affected by color and appearance, which may differ significantly for cross-domain instances.
Thus, to compensate for appearance differences at lower levels, we use style transfer to transform the corresponding regions to a common appearance, as illustrated in Figure \ref{fig:midstyle}.
We use the notation of $C_{A}^{\ell}(P^{\ell},Q^{\ell})$ and $C_{B}^{\ell}(Q^{\ell},P^{\ell})$ to define the denote the transformed features $F^{\ell}_A(P^{\ell})$ and $F^{\ell}_B(Q^{\ell})$.

The similarity metric between two neurons $p\in P^{\ell}$ and $q\in Q^{\ell}$, is defined as the normalized cross-correlation between the common appearance version of their surrounding deep feature patches:
\begin{equation}
	 d(p,q,P^{\ell},Q^{\ell}) = \sum_{i\in N^{\ell}(p), j\in N^{\ell}(q)}\frac{C_A^{\ell}(i;P^{\ell},Q^{\ell})\,C_B^{\ell}(j;Q^{\ell},P^{\ell})}{\|C_A^{\ell}(i;Q^{\ell},P^{\ell})\|\|C_B^{\ell}(j;Q^{\ell},P^{\ell})\|},
\end{equation}
where $N^{\ell}(p)$ and $N^{\ell}(q)$ are the indices of the neighboring neurons of $p$ and $q$, respectively.
$||\cdot||$ denotes the $L_2$ norm throughout the paper.
The neighborhood size is determined by the level: we use $3\times 3$ for the two top levels ($\ell=4,5$) and $5\times 5$ for $\ell=1,2,3$.. 

Transferring the appearance or style of one image to another while keeping the content of the latter has recently been an active research area \cite{gatys2015neural, johnson2016perceptual}.
Various style transfer techniques can be used for our task, as long as they are applied locally on corresponding regions. For instance, the mechanism of Deep Image Analogy \cite{liao2017visual}, which performs a local style transfer can be used to transfer corresponding areas to their middle appearnce, instead of transfering the style of one image to the content of the other and vice versa. In addtion, the idea presented by Huang et al.~\shortcite{huang2017arbitrary} can be adopted. In this work the authors argue that instance normalization performs a form of style transfer by normalizing feature statistics, namely, the style is mainly contained in the mean and the variance of deep features channels. 
This technique greatly simplifies the transfer process and yield plausible results. In our context, however, we do not apply it globally over the entire image, but merely in local regions, which further increases the accuracy of the appearance matching:
\begin{equation}
	C_{A}^{\ell}(P^{\ell},Q^{\ell}) =  \frac{F_{A}^{\ell}(P^{\ell})-\mu_A(P^{\ell})}{\sigma_A(P^{\ell})}\sigma_m(P^{\ell},Q^{\ell}) + \mu_m(P^{\ell},Q^{\ell}) 
\label{eq:midstyle}
\end{equation}
where
\begin{equation}
 \mu_m(P^{\ell},Q^{\ell}) = \frac{\mu_A(P^{\ell})+\mu_B(Q^{\ell})}{2}
\end{equation}
and
\begin{equation}
 \sigma_m(P^{\ell},Q^{\ell}) = \frac{\sigma_A(P^{\ell})+\sigma_B(Q^{\ell})}{2}
\end{equation}
are the mean and standard deviation of the common appearance, and $\mu_A(\cdot),\mu_B(\cdot)\in\mathbb{R}^{K^{\ell}}$  and $\sigma_A(\cdot),\sigma_B(\cdot)\in\mathbb{R}^{K^{\ell}}$ are the spatial mean and  standard deviation over the denoted region, for each of the $K^{\ell}$ channels.
$C_{B}^{\ell}(Q^{\ell},P^{\ell})$ is defined similarly, using $F^{B}_{\ell}(Q^{\ell})$ as the source feature.
The local transfer to a common appearance is illustrated in Figure~\ref{fig:midstyle}.
The advantage of our common appearance metric is demonstrated in Figure~\ref{fig:midstyle_heatmap}: we compute the NBBs between low level features ($\ell=1$) of two corresponding regions, and extracted $k=5$ using the process described in Section~\ref{subsec:kpoints}. As can be seen, the standard cross-correlation in low levels, which is based on color and shape, yields wrong corresponding positions, while our similarity metric localizes the corresponding points better.

\subsection{Meaningful Best Buddies}
\label{subesc:meaningfulBB}

Denote by $ֿ\Lambda^{\ell}$ the candidate NNBs computed in the $\ell$-th level.
At each level, before moving to next one, we filter pairs of best buddies based on their activation values, keeping only pairs where both neurons have strong activation values.
This means that the paired neurons are deemed more significant by the CNN, in the top levels, and indicate significant low-level elements (corners, edges, etc.), in the bottom levels.

In order to be able to compare between activations of neurons from different layers, we compute a \emph{normalized activation map} for each layer $F^{\ell}_A$, which assigns each neuron a value in the range $[0,1]$:
\begin{equation}
	H_{A}^{\ell}(p) = \frac{\|F^{\ell}_A(p)\|-\min_{i}\|F^{\ell}_A(i)\|}{\max_{i}\|F^{\ell}_A(i)\|-\min_{i}\|F^{\ell}_A(i)\|},
\label{eq:normlized_map}
\end{equation}
where $\|F^{\ell}(p)\|$ is the unnormalized activation of a neuron at position $p$ in layer $\ell$. $H_{B}^{\ell}$ is defined similarly with respect to $F^{\ell}_B$.

\begin{figure}
\centering
\includegraphics[width=\linewidth]{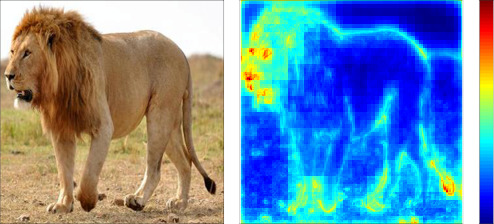} 
\begin{tabular}{cc}
(a) &  \hspace{2.8cm}(b)
\end{tabular}
\caption{Normalized activation map. (a) Original image. (b) The sum of $L=5$ normalized activation maps, each with appropriate upsampling. The NBBs are filtered based on these activations, and thus tend to be located at perceptually important (discriminative, geometrically well-defined) positions.}
 \label{fig:lionresponse}
\end{figure}

Figure~\ref{fig:lionresponse} visualizes the sum of the normalized activation maps over all layers of an image, after upsampling each layer's activation map to the full image resolution. It may be seen that the highly ranked neurons are located in places that are important for the classification of the lion (the face, mane, legs and tail) and geometrically well defined (edges and corners).


Using the normalized activation maps, we seek NBBs which have high activation values, and filter the original set to create:
\begin{equation}
	\tilde{\Lambda}^{\ell} = \left\lbrace (p, q)\in \Lambda^{\ell} \; \bigg\vert \; H_{A}^{\ell}(p)> \gamma \; \textrm{and} \; H_{B}^{\ell}(q)>\gamma \right\rbrace,
\end{equation}
where $\gamma=0.05$ is an empirically determined activation threshold.

Given $\tilde{\Lambda}^{\ell}$, we next refine the filtered NBBs, by generating the finer consecutive set of corresponding search regions, $R^{\ell-1}$, defined by their receptive field in the adjacent layer. In practice, due to observation of Long \etal~\shortcite{long2014convnets}, which showed that similar features tend to respond to similar colors in the centers of their receptive fields, the search area in the finer layer is selected to be half of the receptive field. Then, we perform the search in each pair of corresponding search windows to extract
\begin{equation}
	R^{\ell-1} = \left\lbrace \left( G_{\ell-1}^{\ell}(p^{\ell}_i), G_{\ell-1}^{\ell}(q^{\ell}_i) \right) \right\rbrace_{i=1}^{N^{\ell}},
\label{eq:receptivefield}
\end{equation}
where 
\begin{equation}
G_{\ell-1}^{\ell}(p)=\left[ 2p_x-\frac{r_{\ell-1}^{\ell}}{2}, 2p_x+\frac{r_{\ell-1}^{\ell}}{2} \right] \times \left[  2p_y-\frac{r_{\ell-1}^{\ell}}{2}, 2p_y+\frac{r_{\ell-1}^{\ell}}{2} \right] 
\end{equation}
for a receptive field radius  $r_{\ell-1}^{\ell}$ and central coordinates $p=[p_x,p_y]$, as illustrated in Figure~\ref{fig:bbps} (b). In our case, $r_{\ell-1}^{\ell}$ is equal to $4$ for $\ell=2,3$ and $6$ for $\ell=4,5$.

We repeat this process for the $L=5$ levels.
The final set of correspondences, $\Lambda^0$, between the pixels of the two images $I_A, I_B$, is set to the output of the last activation based filtering, $\tilde{\Lambda}^1$. Note that the lowest layer, $\ell=1$, has the same spatial resolution as the original image.
The entire process is summarized in Algorithm~\ref{alg:crossdomain}.

\begin{algorithm}[t]
\SetAlgoNoLine
\KwIn{Two RGB images: $I_A,I_B$}
\KwOut{A set of corresponding point pairs $\Lambda^0=\{(p_i,q_i)\}_{i=1}^N$}
{\bf Preprocessing:} Extract $\{F_A^{\ell}\}_{\ell=1}^5, \{F_B^{\ell}\}_{\ell=1}^5$ by a feed forward of $I_A, I_B$ through the VGG-19 network.\\
{\bf Initialization:} Set $R^5=\{P^5,Q^5\}$ to the entire domain of $F_A^5$ and $F_B^5$, $C_A^5=F_A^5$ and $C_B^5=F_B^5$ .\\
\For{ $\ell = 5$ to $1$}
{
Extract $\Lambda^{\ell}$ from $C_A^{\ell},C_B^{\ell}$ within corresponding regions $R^{\ell}$.\\
\text{Extract $\tilde{\Lambda}^{\ell}$ by filtering $\Lambda^{\ell}$ based on neural activations}\\
 \If{$\ell>1$ } 
{
Refine the search regions, $R^{\ell-1}=G_{\ell-1}^{\ell}(\tilde{\Lambda}^{\ell})$, using \eqref{eq:receptivefield}
\\
Generate common appearance features $C_A^{\ell-1}, C_B^{\ell-1}$, using \eqref{eq:midstyle}.
}

 }
 
$\Lambda^0=\tilde{\Lambda}^1$ 
\caption{Cross-Domain Deep Correspondence}
\label{alg:crossdomain}
\end{algorithm}

\subsection{Ranking and Spatial Distribution}
\label{subsec:kpoints}

The algorithm as described above produces a (non-fixed) number of NBB pairs.
The corresponding pairs connect points of various quality and reliability, which are not necessarily well distributed across the regions of interest in each image.
To rectify this situation, we refine the selection as described below.

\begin{figure}
\includegraphics[width=\linewidth]{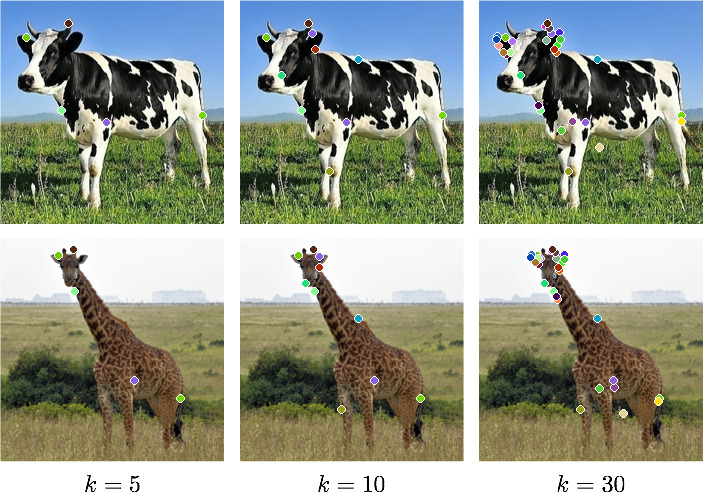}
\caption{Top $k$ spatially scattered NBBs for different values of $k$.}
\label{fig:scale}
\end{figure}

We define the \emph{rank} of a pair $V(p,q)$ as the accumulated activation values of $p$ and $q$ through the levels of the feature hierarchy:
\begin{equation}
	V(p,q) = \sum_{\ell=1}^5 H^{\ell}_{A}(p^{\ell}) + H^{\ell}_{B}(q^{\ell}),
\end{equation}
where $p^{\ell}$ and $q^{\ell}$ are the positions in the $\ell$-th level that eventually led to the selection of $p$ and $q$. 
The top $k$ pairs are unlikely to be well distributed across the images, since most of the highly ranked pairs lay around the same areas, e.g., eyes and mouth in the case of a face.
To extract $k$ highly-ranked and well distributed pairs, we first partition our set of correspondences $\Lambda^0$, into $k$ spatial clusters using the k-means algorithm.
Next, in each cluster we select the pair with the highest rank. The effect of this process is demonstrated in Figure~\ref{fig:scale}.

The coarse-to-fine selection of our algorithm is demonstrated in Figure~\ref{fig:teaser} for $k=5$, highly ranked, spatially scattered points. Each endpoint is annotated with circles that represent the receptive fields of its root points, at layers $\ell=3$ and $\ell=4$, over the original image pixels.

\begin{figure*}
	\centering
	\includegraphics[width=\linewidth]{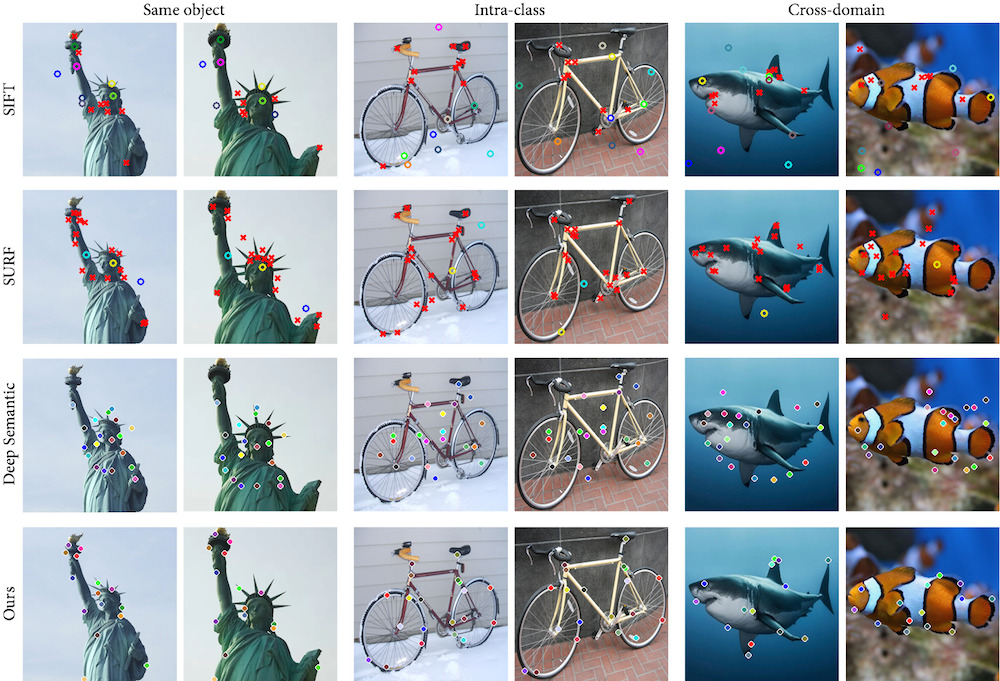}
	\caption{Pairwise key-point localization, comparison to SIFT \cite{lowe2004distinctive}, SURF \cite{bay2006surf} and Deep Semantic Feature Matching \cite{ufer2017deep}. It can be seen that handcrafted descriptors cannot handle difference in appearance. In contrast, \cite{ufer2017deep} matches well intra-class objects, which are semantically identical and normally share the same shape. Our method is designed to overcome large differences in appearance and to handle cross-domain cases.}
	\label{fig:keypoints_compare}
\end{figure*}

\section{Evaluation}
\label{sec:evaluation}

In this section we evaluate the quality of our approach.
It should be noted up front that we know of no other techniques that explicitly aim at cross-domain correspondence.
This also implies that there are no established benchmarks for cross-domain correspondence. We clarify that a cross-domain pair, in this context, means that the images belong to different semantic categories in the ImageNet hierarchy \cite{russakovsky2015imagenet}. 

\ka{Thus, we evaluate our method using several different strategies: (i) we visually compare our NBB-based key point selection with other sparse matching techniques, (ii) we compare our correspondence to state-of-the-art dense correspondence methods, (iii) we perform a user study and compare the wisdom of the crowd statistics to our results, and finally, (iv) we test our method on pairs of objects with different scales, poses and view points to evaluate its robustness to various geometric deformations.}

For every part in this section, many additional examples and results may be found in the supplementary material. 

\begin{figure*}
	\begin{tabular}{ccccccc}
		
		\ Source & SIFT Flow \shortcite{liu2011sift} & Proposal Flow\shortcite{ham2016proposal} & FCSS \shortcite{kim2017fcss} & Deep Analogy \shortcite{liao2017visual} & \textbf{Ours} 
		\\
		\includegraphics[width=2.7cm,trim={0 0 0 0},clip]{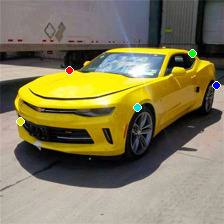}
		&\includegraphics[width=2.7cm,trim={0 0 0 0},clip]{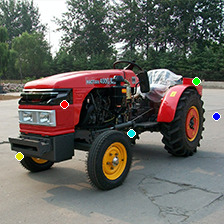}
		&\includegraphics[width=2.7cm,trim={0 0 0 0},clip]{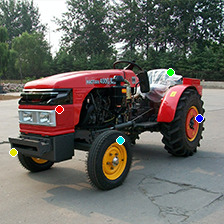}
		&\includegraphics[width=2.7cm,trim={0 0 0 0},clip]{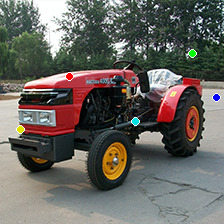}
		&\includegraphics[width=2.7cm,trim={0 0 0 0},clip]{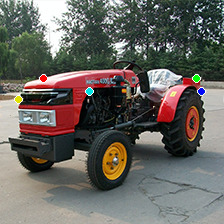}
		&\includegraphics[width=2.7cm,trim={0 0 0 0},clip]{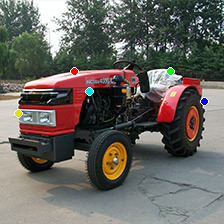}
		\\
		
		\includegraphics[width=2.7cm,trim={0 0 0 0},clip]{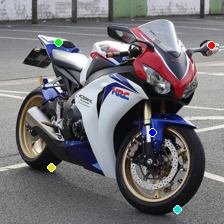}
		&\includegraphics[width=2.7cm,trim={0 0 0 0},clip]{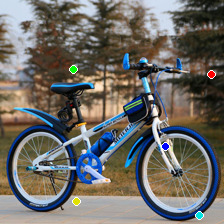}
		&\includegraphics[width=2.7cm,trim={0 0 0 0},clip]{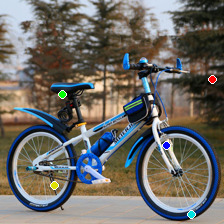}
		&\includegraphics[width=2.7cm,trim={0 0 0 0},clip]{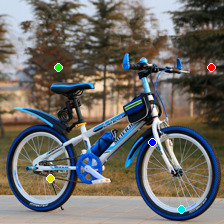}
		&\includegraphics[width=2.7cm,trim={0 0 0 0},clip]{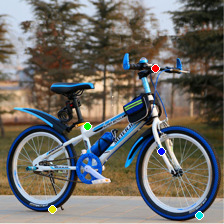}
		&\includegraphics[width=2.7cm,trim={0 0 0 0},clip]{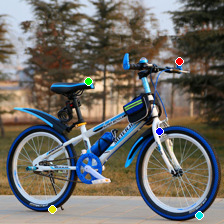}
		\\
		
		\includegraphics[width=2.7cm,trim={0 0 0 0},clip]{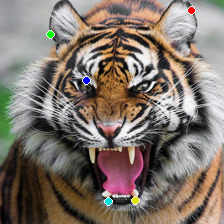}
		&\includegraphics[width=2.7cm,trim={0 0 0 0},clip]{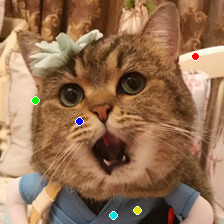}
		&\includegraphics[width=2.7cm,trim={0 0 0 0},clip]{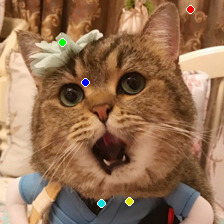}
		&\includegraphics[width=2.7cm,trim={0 0 0 0},clip]{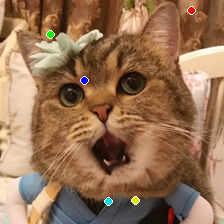}
		&\includegraphics[width=2.7cm,trim={0 0 0 0},clip]{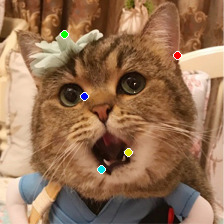}
		&\includegraphics[width=2.7cm,trim={0 0 0 0},clip]{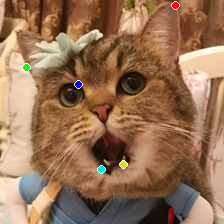}
		\\
		
	\end{tabular}
	\caption{Cross-Domain sparse correspondence, comparison of our method to SIFT Flow \cite{liu2011sift}, Proposal Flow \cite{ham2016proposal}, FCSS \cite{kim2017fcss}, and Deep Image Analogy \cite{liao2017visual}.}
	\label{fig:densecomparison}
\end{figure*}

\subsection{Pairwise Key-Point Localization}
We first visually compare our method to other methods that aim at finding pairs of sparse corresponding key-points.
Normally, these algorithms  first extract points which constitute meaningful candidates for matching in every individual image, then search for their corresponding counterparts in the other image.
Due to the fact that these methods were not explicitly designed to find cross-domain correspondence, we perform only a qualitative comparison, and use image pairs of progressive levels of difficulty: same object, intra-class, and cross-domain.

We compare our method with gradient-based descriptors and a technique based on deep features.
Specifically, we use SIFT \cite{lowe2004distinctive} and SURF \cite{bay2006surf}, by finding the strongest 10 matches using a threshold. In addition, we compare our method with Deep Semantic Feature Matching~\cite{ufer2017deep}, a recent method, which first converts an image space pyramid to a deep features pyramid by transferring each level in the image pyramid to a specific deep layer ({\tt conv\_4} of AlexNet \cite{krizhevsky2012imagenet}). The key points are selected based on the activation and entropy of the features, and the matching is done by minimizing an energy function that considers appearance and geometry.

The results are shown in Figure~\ref{fig:keypoints_compare}.
For the gradient-based descriptors, the matches are marked by circles of corresponding color.
In addition, we indicate with red crosses strong key-points that have no matches. 
As expected, it can be seen that the handcrafted descriptors cannot handle differences in appearance, including intra-class cases.
In addition, it may be seen that deep semantic feature matching \cite{ufer2017deep}, which uses in fact only one layer of deep features, is unable to handle large differences in appearance, which are typical for cross-domain cases. 
The combination of our coarse-to-fine reconstruction and the common appearance transfer enables our approach to identify well-localized semantic similar pairs.
Note, in particular, the highly non-trivial matching pairs between the images of the shark and the clownfish, where the other methods fail completely.

\subsection{Dense Correspondence}
We compare our method to state-of-the-art dense correspondence methods. We separate our evaluation into two parts: cross-domain cases and intra-class pairs.

\paragraph{Cross-Domain}
Dense correspondence methods are typically not designed to handle cross-domain image pairs. Nevertheless, they compute a dense correspondence field, and thus may be used, in principle, to map any point from one image to another.
Figure~\ref{fig:densecomparison} shows several tests designed to test whether by applying several state-of-the-art dense correspondence in this manner, it might be possible to map the key points detected in one image to reasonable matching location in the other.
For these tests, we use several cross-domain image pairs. Each image pair contains objects from different semantic categories, which exhibit drastic differences in appearance, yet feature some semantically similar parts (wheels, eyes, etc.)

The methods used in this evaluation are: SIFT Flow \cite{liu2011sift}, Proposal Flow \cite{ham2016proposal}, FCSS \cite{kim2017fcss}, \ka{and Deep Image Analogy \cite{liao2017visual}}. FCSS is a self-similarity descriptor that is based on a deep network and aims to robustly match points among different instances within the same object class.
Kim et al. \shortcite{kim2017fcss} use a standard flow-based technique to generate a dense correspondence field from this descriptor. \ka{Deep Image Analogy is designed for visual attribute transfer (combining the content of one image with the style of another). It performs the nearest neighbors search, based on PatchMatch \cite{barnes2009patchmatch}, and controlled by parameters which dictate the relative contributions of the content and style sources.} 
Figure~\ref{fig:densecomparison} shows the mapping predicted by these methods for the key points suggested by our approach, and compares them to our result (rightmost column).
With the absence of a ground truth, the results may only be evaluated qualitatively.
Apart from our method and Deep Image Analogy, it is apparent that differences in appearance prevent the other methods to find a reasonable corresponding location for many of the points. While points on wheels of the car and the motorcycle are matched reasonably, this is not the case for other points, such as the tiger's teeth. Also note that many of the points are not mapped onto the foreground object, but rather onto the surrounding background.
\ka{While Deep Image Analogy typically finds correspondences on the foreground object, it may be seen that some of the matches are erroneous or less precisely localized. We attribute these differences to its use of random search and reliance on parameters that should be fine tuned.
This point will be further elaborated in the next section.}

\paragraph{Intra-Class}
In order to quantitatively evaluate our method, we further compare it against an annotated ground truth benchmark of intra-class objects. We use the Pascal 3D+ dataset \cite{xiang2014beyond}, which provides sparse corresponding sets for same class objects with different appearance. For each category, we exhaustively
sample all image pairs from the validation set, and make a comparison between our method to SIFT Flow \cite{liu2011sift}, Proposal Flow \cite{ham2016proposal}, FCSS \cite{kim2017fcss},  \ka{Deep Image Analogy \cite{liao2017visual}} and 3D Cycle Consistency \cite{zhou2016learning}. Cycle consistency is a representative work that considers high-level semantic information for dense matching. Since this method assumes having a shared 3D model of the class, we found it appropriate to include it only in the intra-class comparison. 

Correctness of correspondences is measured by the percentage of correct keypoint transfers (PCK). A transfer is considered correct if the predicted location falls within $\alpha\!\cdot\!\max(H,W)$ pixels from the ground-truth, where $H$ and $W$ are the height and width of the image. We compute the PCK over all the pairs in each category.

Since the benchmark consists of point pairs that may differ from the ones produced by our method, we first extend our sparse correspondence into a dense one. This is done by interpolating our correspondences with Moving Least Squares (MLS) \cite{schaefer2006image} warping field, which is described in more detail in Section~\ref{subsec:imagealignment}.

\begin{figure}
\begin{tabular}{cccc}
\begin{turn}{90}Source\end{turn} 
&\includegraphics[width=0.27\linewidth,height=1.7cm]{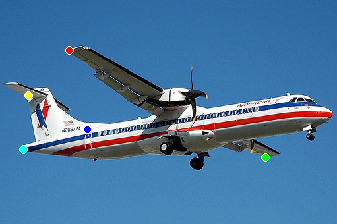}
&\includegraphics[width=0.27\linewidth,height=1.7cm]{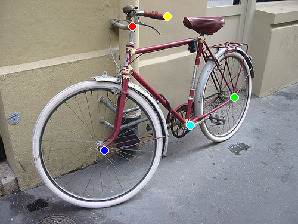}
&\includegraphics[width=0.27\linewidth,height=1.7cm]{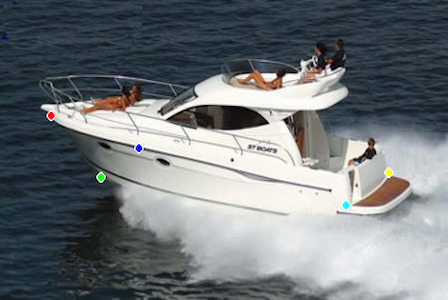}
\\

\begin{turn}{90}SIFT Flow\end{turn} 
&\includegraphics[width=0.27\linewidth,height=1.7cm]{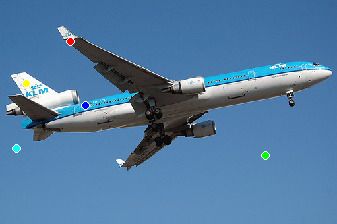}
&\includegraphics[width=0.27\linewidth,height=1.7cm]{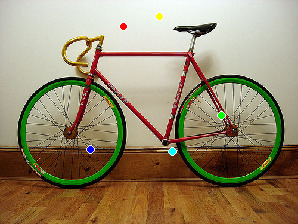}
&\includegraphics[width=0.27\linewidth,height=1.7cm]{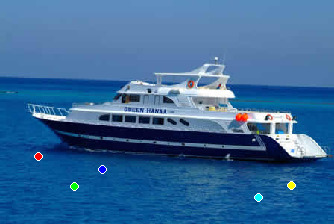}
\\

\begin{turn}{90}Proposal Flow\end{turn} 
&\includegraphics[width=0.27\linewidth,height=1.7cm]{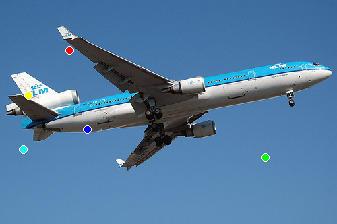}
&\includegraphics[width=0.27\linewidth,height=1.7cm]{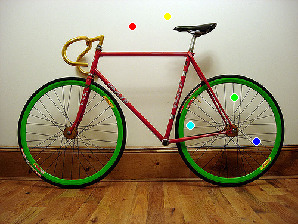}
&\includegraphics[width=0.27\linewidth,height=1.7cm]{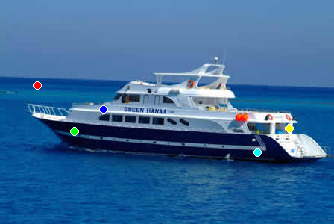}
\\

\begin{turn}{90}FCSS\end{turn} 
&\includegraphics[width=0.27\linewidth,height=1.7cm]{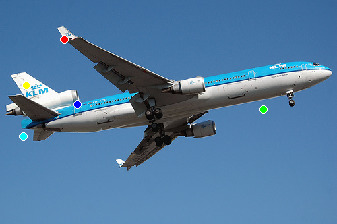}
&\includegraphics[width=0.27\linewidth,height=1.7cm]{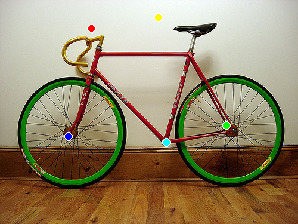}
&\includegraphics[width=0.27\linewidth,height=1.7cm]{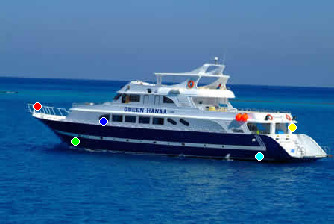}
\\

\begin{turn}{90}3D Cycle\end{turn} 
&\includegraphics[width=0.27\linewidth,height=1.7cm]{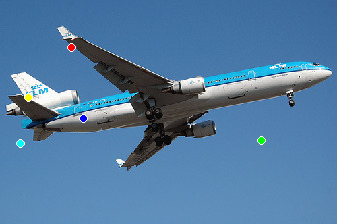}
&\includegraphics[width=0.27\linewidth,height=1.7cm]{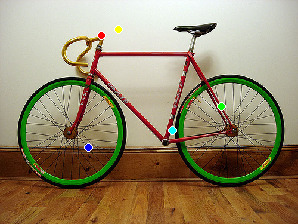}
&\includegraphics[width=0.27\linewidth,height=1.7cm]{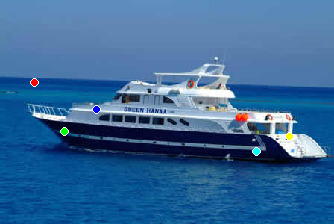}
\\

\begin{turn}{90}Deep Analogy\end{turn} 
&\includegraphics[width=0.27\linewidth,height=1.7cm]{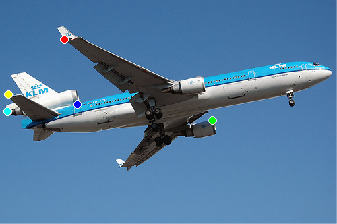}
&\includegraphics[width=0.27\linewidth,height=1.7cm]{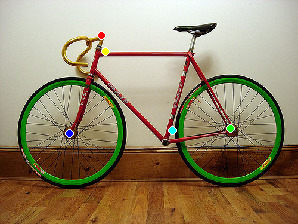}
&\includegraphics[width=0.27\linewidth,height=1.7cm]{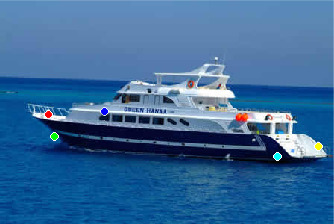}
\\

\begin{turn}{90}Ours\end{turn} 
&\includegraphics[width=0.27\linewidth,height=1.7cm]{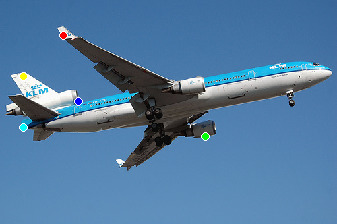}
&\includegraphics[width=0.27\linewidth,height=1.7cm]{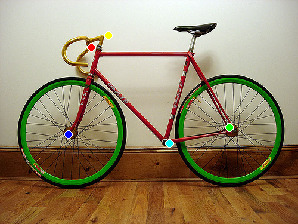}
&\includegraphics[width=0.27\linewidth,height=1.7cm]{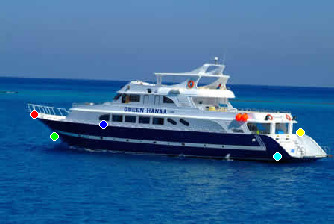}
\\

\begin{turn}{90}Target\end{turn} 
&\includegraphics[width=0.27\linewidth,height=1.7cm]{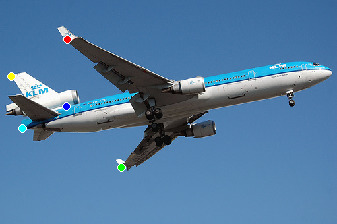}
&\includegraphics[width=0.27\linewidth,height=1.7cm]{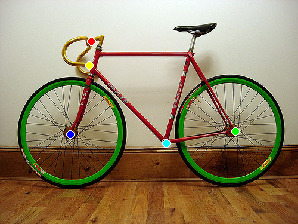}
&\includegraphics[width=0.27\linewidth,height=1.7cm]{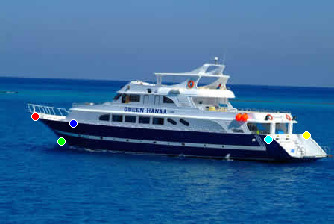}
\\

\end{tabular}
\caption{Same (intra) class correspondence, based on the annotated ground truth of Pascal 3D+ dataset. The comparison was done between our method to SIFT Flow \cite{liu2011sift}, Proposal Flow \cite{ham2016proposal}, FCSS \cite{kim2017fcss}, 3D Cycle Consistency \cite{zhou2016learning}, and Deep Image Analogy \cite{liao2017visual}.}
\label{fig:sameclass}
\end{figure}

The quantitative comparisons between different methods, using PCK with $\alpha=0.1$, are shown in Table~\ref{tab:PCK} and visual comparisons on representative pairs are shown in Figure~\ref{fig:sameclass}.
Our method and deep image analogy, obtain better performance than methods based on low-level features, e.g., SIFT Flow. In addition, our method performs better than cycle consistency (often by a significant margin), even though the features we use were not trained on the Pascal 3D+ dataset. 

%

\setlength{\tabcolsep}{6pt}
\renewcommand{\arraystretch}{1}
\begin{table*}[t]
\begin{center}
\vspace{4pt}
\resizebox{\textwidth}{!}{
\begin{tabular}{ ccccccccccccc|c }
\toprule
& \small{aeroplane} & \small{bicycle} & \small{boat} & \small{bottle} &\small{bus} & \small{car} & \small{chair} & \small{table} &\small{motorbike} & \small{sofa} & \small{train} & \small{monitor}& \textbf{mean}\\
\midrule
\small{SIFT Flow}
& 10.3 & 10.9  & 3.4 & 23.5 & 13.0 & 13.4 & 8.2 & 5.2 & 9.1 & 15.3 & 13.6 & 22.1 & 12.3\\

\small{Proposal Flow}
& 14.3 & 9.7  & 12.4 & 38.6 & 9.1 & 14.2 & 21.4 & 15.3 & 10.7 & 23.9 & 6.5 & 25.1 &16.7\\
\small{FCSS}
& 20.7 & 22.2  & 11.8 & 50.0 & 22.5 & 31.3 & 18.2 & 14.2 & 15 & 28.9 & 8.7 & 30 & 22.8\\

\small{3D Cycle}
& 20.8 & 25.0  & 4.9 & \textbf{51.0} & 21.1 & \textbf{33.8} & 37.9 & 11.9 & 15.0 & 32.2 & 13.7 & 33.8 & 25.0\\

\small{Deep Image Analogy}
& 20.0 & 34.2  & 7.9 & 44.3 & 18.2 & \textbf{33.8} & \textbf{38.7} & 10.9 & 17.0 & 38.4 & 14.7 & 35.8 & 28.3\\

\small{Ours}
&  \textbf{21.4} &  \textbf{38.8}  &  \textbf{14.0} & 48.2 &  \textbf{32.5} &  33.2 & 26.3 &  \textbf{20.0} &  \textbf{23.4} &  \textbf{48.4} &  \textbf{24.8} & \textbf{45.8} & \textbf{31.4}\\

\bottomrule
\end{tabular}
}
\end{center}
\caption{Correspondence accuracy for intra-class objects, measured in PCK ($\alpha=0.1$). The test is conducted on the validation pairs of each category of the PASCAL 3D+ dataset.
\vspace{-5mm}
}
\label{tab:PCK}
\end{table*}

\subsection{User evaluation}
In the case of same-scene images or intra-class objects,  it is pretty straightforward for humans to annotate semantically similar points.
However, in case of cross-domain instances, the matches are more ambiguous, and there might be no obvious or unique pixel-to-pixel matches.
To evaluate the matching obtained by NBB, we conducted a user study. Each of the 30 study participants was shown 5 image pairs, which were chosen randomly each time from a larger set of 13 cross-domain pairs.
Each participant was presented with $k = 5$ points in the first image, and was asked to indicate $k$ best corresponding points in the second image.
The $k$ points were automatically extracted by our method, as described in Section~\ref{subsec:kpoints}.

\begin{figure}
\centering
\includegraphics[width=\linewidth]{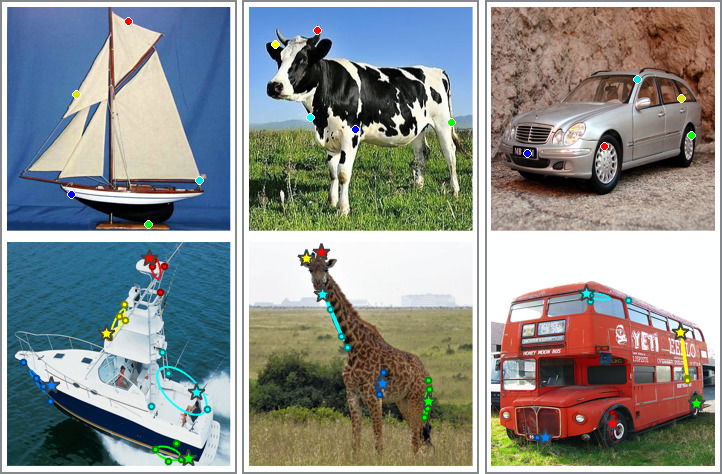} 
\begin{tabular}{ccc}
(a) &  \hspace{2.1cm}(b) &  \hspace{2.1cm}(c) 
\end{tabular}
\caption{\ka{Comparison of our results to human annotation. The top row shows three of the images that were presented to the users, with our top $k=5$ NNBs. The bottom row shows the corresponding images that users were asked to annotate. For each point in the top image, a Gaussian distribution of the users' annotations is visualized in corresponding image as an ellipse of matching color, along with 4 representative user annotations (dots) and our NBB counterparts (stars).}}
 \label{fig:usertest}
\end{figure}

\ka{In order to quantitatively evaluate our algorithm based on the collected user annotations, we measure how well our results, as well as those of other algorithms, are aligned with these annotations. Specifically, for each individual point, we define the similarity measurement between that point and the user annotations to be the value of the probability density function at the point, where the density function is a 2D Gaussian distribution, whose parameters are estimated using maximum likelihood. Thus, when there is a strong consensus among the users regarding the position of the corresponding point, any significant deviation will be assigned a low score, unlike the case where the variance between the different users is large. For each algorithm we averaged the score of all the points. The results are shown in Table~\ref{tab:user_study}.
	
For qualitative evaluation, three results are demonstrated in Figure~\ref{fig:usertest}.
The first image in each pair (first row) is the one that was presented to the users with 5 points (dots of different color). On the second image in each pair (second row), we visualize the Gaussian distribution of the users' responses using ellipses of corresponding color. We also performed k-means clustering of the responses for each color into four clusters,
and show the centers of the clusters as colored dots.
The corresponding points found by our approach are marked as stars on the same image.}

Note that the corresponding points indicated by different users exhibit considerable variance, due to the semantic uncertainty of their true location. This is especially evident in Fig.~\ref{fig:usertest}(a), where both the structures of the sailboat and the yacht and the viewpoint differ greatly. Nevertheless, it can be seen that the matches produced by our approach are well aligned with the distribution of human annotations.

\begin{table}
\begin{center}
\vspace{4pt}

\begin{tabular}{cccccc}
\toprule
&  \small{SIFT}& \small{Proposal} & \small{FCSS} & \small{Deep Image} &  \small{Ours} \\
 & \small{ Flow}& \small{Flow} & & \small{ Analogy} &  \\
\midrule
\small{Score}
&0.032 & 0.042 & 0.084 & 0.123 & \bf{0.163} \\
\bottomrule
\end{tabular}

\end{center}
\caption{\ka{Quantitative evaluation of the user study. The value represents the average probability of the corresponding point to be part of the user statistics, whose density function is modeled by a 2D Gaussian distribution.}
\vspace{-5mm}
}
\label{tab:user_study}
\end{table}

\subsection{Robustness and Limitations}
%
 
\begin{figure}
\centering
\includegraphics[width=\linewidth]{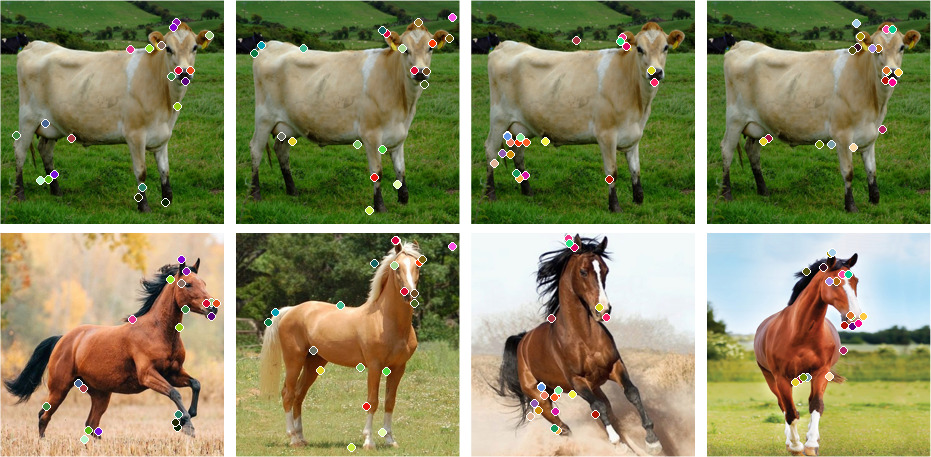} 
\caption{NBB robustness to pose difference for cross-domain objects. Each column presents a pair of cow (static) and a horse (in different poses) with its top $k=20$ NBBs.
}
 \label{fig:pose}
\end{figure}

\paragraph{Pose and Scale}
In order to test the robustness of our method to pose differences, we match the same cow image to four images of horses, each featuring a different pose. The results are shown in Figure~\ref{fig:pose}. It may be seen that our method can cope quite well with these cross-domain objects, despite the different poses.

As for robustness to scale differences: although we utilize a hierarchy of features, our method searches for best-buddies only between features at the same level, which might be interpreted as sensitivity to scale. However, since the classification network is trained for scale invariance, this invariance is reflected in its deep features. For instance, Figure~\ref{fig:scale} demonstrates correct matches between the giraffe and the cow heads despite their different scale. A stress test demonstrating robustness to scale differences is included in the supplementary material.

\begin{figure}
\centering
\includegraphics[width=\linewidth]{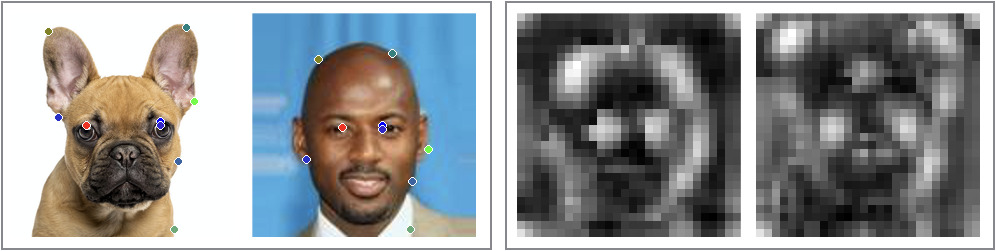} 
\begin{tabular}{cc}
(a) &  \hspace{3.4cm}(b)
\end{tabular}
\caption{Erroneous semantic correspondence due to geometric similarity. (a) Incorrect matches between the ears of the dog and the bald head of the man, and absence of matches between the semantically similar nose and mouth areas. (b) correlation between each cell in the $\ell=4$ layer of one image and its nearest neighbor in the other corresponding map. }
 \label{fig:limitation}
\end{figure}

\paragraph{Erroneous matches}
We next show that the information which is encoded in the high-level features is not purely semantic and geometry might also influence similarity of deep patches. This might lead to erroneous semantic correspondence and missing matches. For example, Figure~\ref{fig:limitation} demonstrates such a failure case where the round silhouette of the man's bald head shares highly correlated features with the edges of the dog's ears, leading to semantically incorrect corresponding pairs. In addition, due to the significant geometry differences, between their nose and mouth, the network is unable to localize shared points around these semantically similar areas, which is a challenging task even for a human. Figure~\ref{fig:limitation} (b) shows the correlation between each cell in the $\ell=4$ layer of one image and its nearest neighbor in the other corresponding map. It can be seen that high correlation exists around the bald head and the dog's ears, while the correlation around the nose and mouth is relatively low.

\section{Applications}
\label{sec:apps}


Having computed a sparse set of correspondences between a pair of cross domain images, it is possible to warp the two images, such that their matching content becomes aligned, as well as to define a dense correspondence field between the two images. This, in turn, enables a variety
of graphics applications, some of which are discussed below.

\ka{We note that the warping operations are performed in the pixel domain. The deep features extracted by the VGG network are intentionally translation invariant; thus,
applying non-translational transformations directly onto the deep feature maps could lead to distortions in the reconstructed image.}

\subsection{Cross-Domain Image Alignment}
\label{subsec:imagealignment}


Given a sparse set of NBBs for a pair of cross-domain images, our goal is to define a continuous warping field that would align the NBB pairs, while distorting the rest of the image as little as possible.
Rather than applying optical flow or Nearest-Neighbor Field methods, we employ the moving least squares (MLS) image deformation method \cite{schaefer2006image} for this purpose.
More specifically, given a set of matching pairs $\{(\alpha_i \in I_A, \beta_i \in I_B)\}$, we compute a set of midpoints $\{\eta_i = 0.5(\alpha_i + \beta_i)\}$. The two images are then aligned by warping each of them such that its set of matched points is aligned with $\{\eta_i\}$. 



\begin{figure*}
	\includegraphics[width=0.9\linewidth]{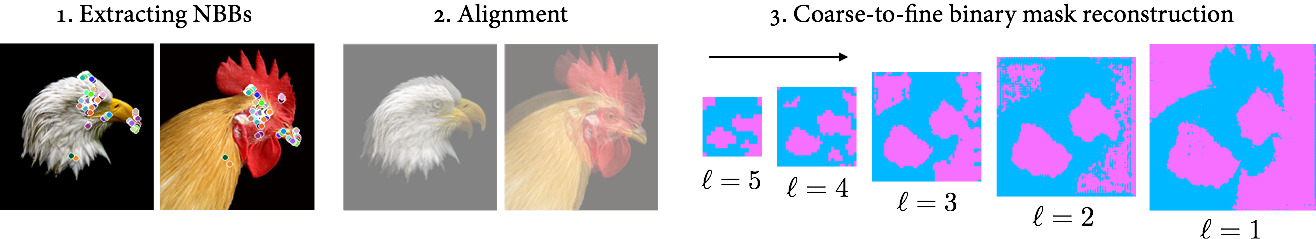}\\ 
	\centering
	(a) \\\vspace{0.5cm}
	\includegraphics[width=0.9\linewidth]{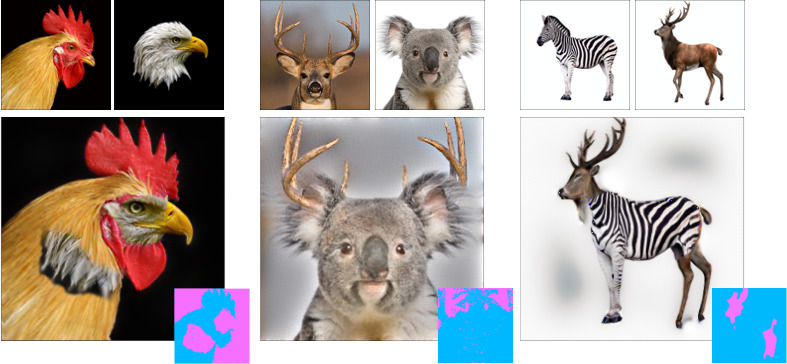}\\ (b)
	\caption{Semantic hybridization. (a) The two input images are first aligned based (only) on their shares parts using the NBBs. Then, a low-resolution semantic mask ($\ell=5$) is selected based on neural activations. The mask is then propagated, coarse-to-fine, into the original image resolution. (b) Resulting hybrids: original images (on top), final binary mask (down), resulting hybrid (middle).}
	\label{fig:hybrid}
\end{figure*}

\subsection{Semantic Hybridization}
\label{sec:hybridization}

A hybrid image is one formed by combining parts from two images of different subjects, such that both subjects contribute to the result. Admittedly, this is not a well defined task;
the hybridization strategy can be dictated by an artist, who selects which segments or attributes should be selected from each image, which segments to blend, and how to combine the colors \cite{benning2017nonlinear}.
As discussed below, aligning cross-domain images can be instrumental in this process, but it also makes it possible to create some interesting image hybrids automatically.

Given two cross-domain images, we aim to include in the hybrid their most semantically important and distinctive parts.
This is done by first aligning the two images, as described above, and then combining their deep feature maps in a coarse-to-fine fashion.

Specifically, given two aligned images $\tilde{I}_A$ and $\tilde{I}_B$, the goal is to generate a binary mask that indicates which areas of each input image are considered to be more meaningful and discriminative, and thus should be included in the hybrid.

We begin by forming such a mask for the coarsest level ($L=5$) of the deep features pyramid, defined as described in Section~\ref{subsec:deepfeatures}. The two feature tensors corresponding to the two aligned images $F^L_A$ and $F^L_B$ are normalized, and a hybrid feature tensor $F^L_H$ is formed by taking the most strongly activated neuron at each spatial location:
\begin{equation}
	F^{L}_H(p) = \begin{cases}
F^{L}_A(p) & \frac{\|F^{L}_A(p)\|}{\sum_i\|F^{L}_A(i)\|} \geq \frac{\|F^{L}_B(p)\|}{\sum_i\|F^{L}_B(i)\|}\\
F^{L}_B(p) & \text{otherwise}.
\end{cases}
\end{equation}

The hybrid feature map is then propagated down the hierarchy. To obtain the map $F^\ell_H$ from $F^{\ell+1}_H$, we first use feature inversion \cite{mahendran2015understanding}, without regularization, to obtain an inverted feature map $\bar{F}^{\ell}_H$, which is then projected onto the space of hybrids of $F^{\ell}_A$ and $F^{\ell}_B$. Specifically,
\begin{equation}
F^{\ell}_H(p) = \begin{cases}
F^{\ell}_A(p) & \left\Vert {F}_A^{\ell}(p) - \bar{F}_H^{\ell}(p) \right\Vert  \leq  \left\Vert {F}_B^{\ell}(p) - \bar{F}_H^{\ell}(p) \right\Vert,\\
F^{\ell}_B(p) & \text{otherwise}.
\end{cases}
\end{equation}

When reaching the pixel level ($\ell = 0$), we do not perform the projection above, and use the inverted feature map $\bar{F}^{0}_H$ as the resulting image hybrid, since this produces slightly softer transitions across boundaries between regions from different images.

Figure \ref{fig:hybrid}(a) demonstrates the process described above, and the final hybrid shown in Figure \ref{fig:hybrid}(b), along with two additional examples. It can be seen that in each of the examples the unique, discriminative parts of the animal, were selected automatically, i.e., the comb of the rooster, the eagle's beak, the deer's antlers.

Our cross-domain alignment may also be used by an interactive tool that enables artists to effortlessly produce hybrid images.

\begin{figure*}
	\begin{tabular}{c}
		\includegraphics[width=0.9\linewidth]{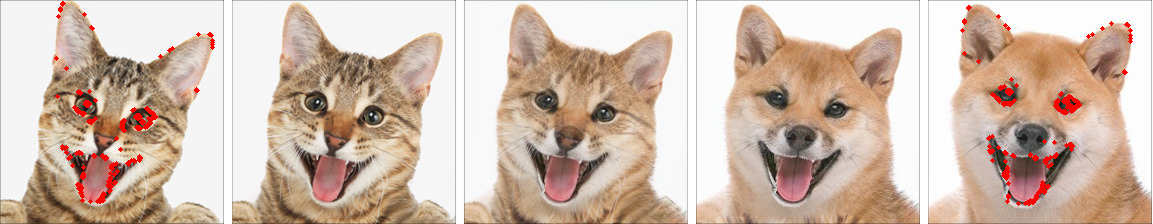}\\
		\includegraphics[width=0.9\linewidth]{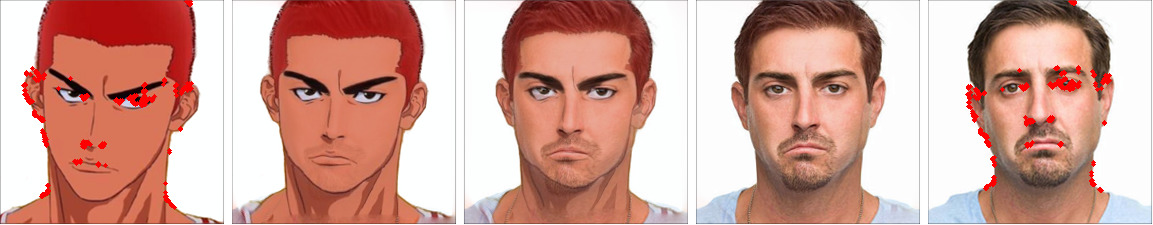}\\
		\includegraphics[width=0.9\linewidth]{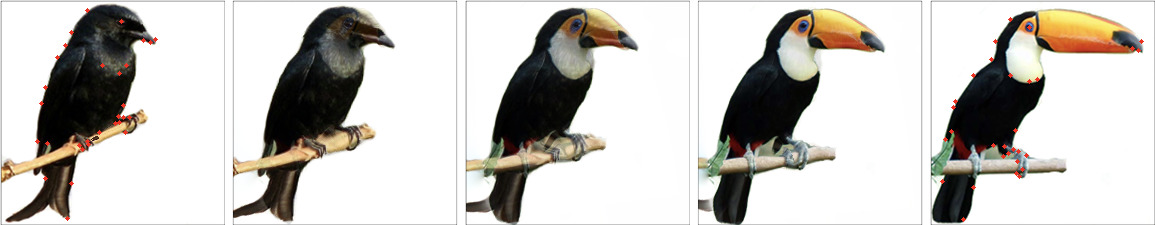}
	\end{tabular}
	\caption{Fully automated image morph. The two original images (left and right columns) are warped into a morph sequence ($25\%, 50\%, 75\%$) based on our sparse NBBs (red points), which replace the user provided correspondences in the energy function of \cite{liao2014automating}.}
	\label{fig:morph}
\end{figure*}

\subsection{Automatic Cross-Domain Image Morphing}
\label{subsec:morph}
The process of image morphing is based on dense correspondence with well defined motion paths. Typically, the motion paths are first defined between sparse corresponding features and then interpolated into dense smooth trajectories. Based on these trajectories, the images are warped and blended, to produce an animation. Since in many cases, an effective morph requires a semantic understanding of the image content, creating this mapping usually involves significant user interaction, using tagged features such as points, lines, curves, or grids \cite{wolberg1998image}.
 
Below, we utilize our sparse NBBs to modify the semi-automated image morphing method of Liao \etal~\shortcite{liao2014automating} into a fully automated method for cross-domain images.
Liao et al.~generate a dense correspondence between two images by minimizing the following energy functional:
\begin{equation}
	E = E_{\mathrm{SIM}}+E_{\mathrm{TPS}}+E_{\mathrm{CORR}},
\label{eq:morphenergy}
\end{equation}
where the first term is the SSIM index \cite{wang2004image}, which is responsible to rank the distances between pairs of patches, the second is the thin-plate spline (TPS) smoothness term,
while the last term ensures that the warping field follow the set of sparse correspondences.
After finding a dense correspondence, pixel trajectories are defined using quadratic motion paths. We refer the reader to \cite{liao2014automating} for more details.

In the original method of Liao et al. \shortcite{liao2014automating}, the third term is provided by the user, by manual marking of corresponding points. This part is crucial for the success of the morph, especially in the case of cross-domain instances. In our implementation, these manual correspondences are replaced with our automatically determined NBBs.
Figure~\ref{fig:morph} shows the resulting morph on three cross-domain examples.
These results are fully automatic.

\begin{figure}
\centering
\includegraphics[width=\linewidth]{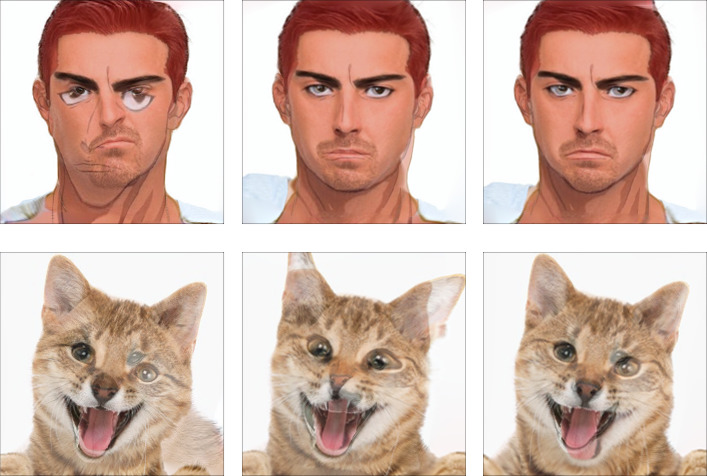}
\parbox{0.33\linewidth}{\centering (a)}~%
\parbox{0.33\linewidth}{\centering (b)}~%
\parbox{0.33\linewidth}{\centering (c)}~%
\caption{Image morphing comparison. The middle image in a morph sequence generated using the method of Liao et al.~\shortcite{liao2014automating} with different sets of sparse correspondences: (a) No $E_{\mathrm{CORR}}$ term. (b) Using corresponding facial landmarks detected by \cite{zhu2015face} (c) Using corresponding facial landmarks detected by \cite{kowalski2017deep}.}
\label{fig:face_landmarks}
\end{figure}

\paragraph{Comparison.}
Our method enables alignment between similar parts of human and animal faces, as can be seen in various examples in the paper and in the supplementary material.
Using the morphing application, we next compare our face alignment with two methods designed specifically for automatic extraction of facial landmarks. The first method is geometry based \cite{zhu2015face} and the other is deep learning based \cite{kowalski2017deep}.
Both methods consist of two steps: face region detection and landmark extraction.
It should be noted that these methods are not intended for animal faces and, indeed, these methods fail in the face detection stage. After manually marking the animals' face regions, the methods were able to extract some landmarks, which we used as corresponding points to replace our NBBs in the $E_{\mathrm{CORR}}$ term of Eq.~\eqref{eq:morphenergy}. We present the middle image of the resulting sequences in Figure~\ref{fig:face_landmarks}.

%

\section{Conclusions and future work}

We have presented a technique for sparse correspondence between cross-domain images. The technique is based on computation of best buddies pairs between the deep feature maps of a pre-trained classification CNN. The performance of our method goes a step beyond state-of-the-art correspondence methods, allowing to establish correspondence between images that do not share common appearance, and their semantic point correspondences are not necessarily obvious by considering local geometry.

Our automatic correspondences are surprisingly very close to those made manually by humans. We attribute this to the following factors: (i) the analysis is based on a pre-trained network, whose features at the deeper layers exhibit different kinds of invariance and were trained for semantic classification; (ii) our hierarchical analysis focuses only on regions that were indicated to have high-level similarity; (iii) as we go down the hierarchy, we search for best buddies in rather small regions, which are easier to bring to a common ground, thereby facilitating local matching; and (iv) the search focuses only on highly active regions, avoiding spurious matches.

Encouraged by our results, we are considering a number of avenues to continue exploring the use of neural best buddies. One avenue is considering a set of images, and applying co-analysis and co-segmentation, where the best buddies are mutual across the set. Since the neural best buddies are located in regions that are active in classification, we believe that they can also facilitate in foreground extraction. Another research avenue is to further advance the hybrid application. We believe that the combination of sparse cross-domain correspondence together with a deep analysis of the activation maps, can lead to creation of coherent hybrids of different objects. Lastly, we would also like to consider developing neural best buddies over pre-trained networks, which were trained for tasks other than classification.

\section{Acknowledgments}
\label{sec:ack}
We thank the anonymous reviewers for their helpful comments.
This work is supported by National 973 Program (No.2015CB352500) of China and the ISF-NSFC Joint Research Program (2217/15, 2472/17).

\bibliographystyle{ACM-Reference-Format}
\bibliography{references} 

\end{document}


\title{\Huge \textbf{Supplemental Material for ``Neural Best-Buddies: Sparse Cross-Domain Correspondence"}}
\date{paper 313}

\maketitle
\tableofcontents
\newpage
\input	{supp_user_study}
\input	{supp_key_points}
\input	{supp_dense}
\input	{supp_limitation}
\section{Applications}
More examples for results of the applications mentioned in the paper.
\subsection{Automatic Cross-Domain Image Morphing}
\begin{figure}[h!]
\centering
\includegraphics[width=\linewidth]{figures/supplementary/morphing}
\caption{Fully automated image morph. The two original images (left and right columns) are warped into a morph sequence ($25\%, 50\%, 75\%$) based on our sparse NBBs (red points), which replace the user provided correspondences in the energy function of \cite{liao2014automating}.}
 \label{fig:morphing}
\end{figure}

\newpage
\subsection{Semantic hybrids}
\begin{figure}[h!]
\centering
\includegraphics[width=0.9\linewidth]{figures/supplementary/hybrids_explanation}\\ (a) \\
\includegraphics[width=0.9\linewidth]{figures/supplementary/hybrids}\\ (b)
\caption{Semantic hybridization. (a) The two input images are first aligned based (only) on their shares parts using the NBBs. Then, a low-resolution semantic mask ($\ell=5$) is selected based on neural activations. The mask is then propagated, coarse-to-fine, into the original image resolution. (b) Resulting hybrids: original images (on top), final binary mask (down), resulting hybrid (middle).}
 \label{fig:hybrids}
\end{figure}

\newpage
\subsection{Transfusive Image Manipulation}

\begin{figure}[h!]
\centering
\includegraphics[width=\linewidth]{figures/supplementary/transfusive}
\caption{Transfusive image manipulation based on our NBB-based image alignment. Brush strokes painted on the left image are automatically transferred to the right image.}
 \label{fig:transfusive}
\end{figure}

\newpage

\bibliographystyle{plain}
\bibliography{references}